\documentclass[10pt, a4paper]{article}

\usepackage[final]{lrec2026} % this is the new style
\usepackage{booktabs}
\usepackage{tabularx}
\usepackage{adjustbox}
\usepackage[dvipsnames]{xcolor}
\usepackage{multirow}
\usepackage{pifont}
\usepackage{float}
\usepackage{algorithm}
\usepackage{algorithmic}
\usepackage{enumitem}
\usepackage{newfloat}
\usepackage{listings}
\usepackage[most]{tcolorbox}
\usepackage{amsmath}
\usepackage[bottom]{footmisc}
\usepackage{makecell}
\title{MUStReason: A Benchmark for Diagnosing Pragmatic Reasoning in Video-LMs for Multimodal Sarcasm Detection.}

\name{Anisha Saha\textsuperscript{\rm 1, \rm 2}, Varsha Suresh\textsuperscript{\rm 2}, Timothy Hospedales\textsuperscript{\rm 3, \rm 4}, Vera Demberg\textsuperscript{\rm 1, \rm 2}} 

\address{\textsuperscript{\rm 1}Max Planck Institute for Informatics, Saarland Informatics Campus, \textsuperscript{\rm 2}Saarland University, \\
         \textsuperscript{\rm 3}The University of Edinburgh, \textsuperscript{\rm 4}Samsung AI Center, Cambridge \\
         ansaha@mpi-inf.mpg.de,
         vsuresh@lst.uni-saarland.de, \\ t.hospedales@ed.ac.uk, vera@lst.uni-saarland.de\\}

\abstract{
Sarcasm is a specific type of irony which involves discerning \textit{what is said} from \textit{what is meant}. 
Detecting sarcasm depends not only on the literal content of an utterance but also on non-verbal cues such as speaker's tonality, facial expressions and conversational context. 
However, current multimodal models struggle with complex tasks like sarcasm detection, which require identifying relevant cues across modalities and pragmatically reasoning over them to infer the speaker’s intention.
To explore these limitations in VideoLMs, we introduce \textbf{MUStReason}, a diagnostic benchmark enriched with annotations of modality-specific relevant cues and underlying reasoning steps to identify sarcastic intent. In addition to  benchmarking sarcasm classification performance in VideoLMs, using MUStReason we quantitatively and qualitatively  evaluate the generated reasoning by disentangling the problem into perception and reasoning and aim to pinpoint the current gaps in these VideoLMs. Furthermore, to facilitate structured pragmatic reasoning, we propose PragCoT, a framework that steers VideoLMs to focus on implied intentions over literal meaning, a property core to detecting sarcasm.
 \\ \newline  }

\begin{document}

\maketitleabstract

\section{Introduction}

Conversations in day-to-day life often involve the use of rhetorical devices like humor, irony, and sarcasm. These forms of expression have an underlying intent that is opposite to their literal meaning. Some forms of sarcasm, such as \textit{“Never? Isn't that usually when you go to the gym?”}, is explicitly conveyed through the speaker’s utterance and can be directly identified from the text. In contrast, others like \textit{“It was her, right?”} rely on additional contextual or multimodal cues like speaker's tonality to be understood as sarcastic \cite{caucci2012social}. 

Research has shown that tone of voice \citep{rockwell2000lower, cheang2008sound, woodland2011context}, facial micro-expressions \cite{Mishra_Kanojia_Bhattacharyya_2016}, and the broader temporal or situational context provide essential information for disambiguating sarcasm. This task is challenging even for humans, as it requires reasoning over cues from diverse sources \cite{farha2023sarcasm, farabi2024survey}. Thus, sarcasm detection becomes a multimodal reasoning challenge which relies on perceiving and integrating subtle cues across modalities.

%Recent advances in multimodal learning have led to models that integrate information from vision, audio, and text through cross-modal fusion and alignment techniques \cite{yin2024survey,xu2023multimodal}. 
Previous work \cite{10.24963/ijcai.2024/887} has shown that for multimodal models, performing complex tasks such as sarcasm detection from conversational videos is challenging because unlike traditional multimodal tasks such as captioning or visual question answering, sarcasm detection relies on inferences about underlying intent rather than simple integration of factual information. Thus, interpreting sarcasm requires pragmatic reasoning in order to resolve the incongruity between the literal meaning of an utterance and non-verbal cues. %While traditional multimodal reasoning builds a rationale by integrating multiple surface-level cues, pragmatic reasoning derives the subtle intentions entangled across modalities.
%%
% First, these models are shown to often fail in instances where perceiving non-verbal cues are essential for interpretation \cite{bhosale2023sarcasmsightsoundbenchmarking}.
% %%
%  Secondly, although these models achieve strong performance on perception-oriented tasks such as retrieval, classification, and captioning, they frequently struggle with tasks requiring higher-order inferential reasoning \cite{li2025perception,xu2024llava,zhang2024improve}.
% %%
% This limitation stems in part from training objectives that prioritize surface-level pattern recognition over the deeper inferential reasoning required for tasks such as sarcasm detection \cite{li2025perception}.
% %%

Most available datasets provide only coarse-grained binary labels and lack annotations that localize relevant multimodal cues or outline the reasoning process involved, revealing the gap in existing resources for evaluating how multimodal models reason about sarcasm. To bridge this gap, we propose \textbf{MUStReason} which provides fine-grained annotations useful for investigating reasoning about sarcasm in Video-Language Models (VideoLMs). This work is the first to introduce a diagnostic benchmark which provides access to reasoning-aligned annotations enabling recognition of key areas where models fail (perceptual issues or faulty reasoning) and provides an in-depth evaluation of VideoLMs' performance in detecting sarcasm. In addition, we propose PragCoT, a pragmatic reasoning framework, which enables derivation of implied intentions from conversational videos, a key step towards detecting sarcasm. Our key contributions are: 
\begin{itemize}
    \item  We introduce MUStReason to enable investigation of perceptual and reasoning gaps in VideoLMs while identifying sarcasm.
    \item We introduce PragCoT, a pragmatic reasoning framework to interpret the literal meaning of the combined modalities as well as underlying intent conveyed by contextual cues. 
    \item  We benchmark sarcasm classification in VideoLMs with and without structured reasoning.

\end{itemize}
% \textbf{C1: Benchmarking sarcasm detection in VideoLMs}: With built-in capabilities for interpreting diverse modalities, we investigate how well VideoLMs perform in classifying sarcastic videos and whether reasoning over multimodal cues further boost their classification ability. \textbf{C2: A sarcasm reasoning dataset to identify challenges in VideoLMs}:  Using MUStReason, we investigate whether the main challenge for VideoLMs in determining sarcasm lies in perception or reasoning. We further investigate whether LLMs outperform VideoLMs in sarcasm detection when provided with accurate perceptual cues. \textbf{C3: A framework for structured reasoning about sarcasm}:  Pragmatic reasoning adds a layer of complexity to multimodal reasoning as it requires interpreting the underlying intent conveyed by contextual cues and the literal meaning of the combined modalities. To enable pragmatic reasoning in VideoLMs, we propose the PragCoT framework, which reasons about sarcasm via detailed and grounded explanation.  

% \anisha{Add conclusions after the experiments section is finalized}

\section{Related Work}
\subsection{Multimodal Sarcasm Detection }
Sarcasm is a complex rhetorical device where the intended meaning contrasts with the literal surface form. 
According to Gricean implicature theory \cite{grice1975logic}, sarcasm emerges when a speaker violates conversational norms, prompting listeners to infer alternative intent. 
For instance, the utterance \textit{``I love the idea!"} may be interpreted as sarcastic when non-verbal cues such as facial expressions and tonality of the speaker suggest the opposite meaning \cite{attardo2000irony}. 
Inferring sarcasm requires perceiving and reasoning over these non-verbal cues \cite{farabi2024survey,caucci2012social}.
%% 
%Specifically, research highlights the crucial role of non-verbal markers in identifying sarcasm. 
Studies show that over 76\% of ironic utterances in face-to-face settings include explicit verbal, paraverbal, or non-verbal markers \cite{athanasiadou2020diversity}. Facial expressions and gestures help humans to detect sarcasm \cite{giustolisi2021role}, with indicators such as eye-rolls \cite{Tabacaru_Lemmens_2014} and mouth movements \cite{rockwell2000lower}. Prosodic features and vocal cues like slower tempo and lower pitch also contribute significantly \cite{cheang2008sound}. 
%These findings underscore that recognizing sarcasm requires perceiving relevant modalities and reasoning over them.

Early computational approaches focused on textual sarcasm detection in social media \cite{hazarika-etal-2018-cascade, poria-etal-2016-deeper}, but later works recognized the importance of non-verbal context. MUStARD \cite{castro-etal-2019-towards} introduced a video-based sarcasm dataset, followed by MUStARD++ \cite{ray2022multimodal}, which incorporated emotion and modality-based sarcasm categories. These datasets enabled models to leverage facial, auditory, and contextual signals, prompting the development of multimodal fusion techniques using attention \cite{pramanick2022multimodal, aggarwal2023multimodal}, contrastive learning \cite{zhang2021multi}, and adaptations of transformer-based architectures \cite{pan-etal-2020-modeling, babanejad2020affective}. However, current models often struggle with sarcasm detection for cases that rely primarily on non-verbal cues \cite{bhosale2023sarcasmsightsoundbenchmarking}, partly due to datasets using only binary sarcasm labels without detailed annotations of relevant cues. This hinders error analysis and targeted improvements. To address this gap, we introduce fine-grained multimodal annotations that describes cues relevant for sarcasm and how to reason over them to detect sarcasm.
\begin{figure*}[t!]
    \centering    \includegraphics[width=0.8\linewidth]{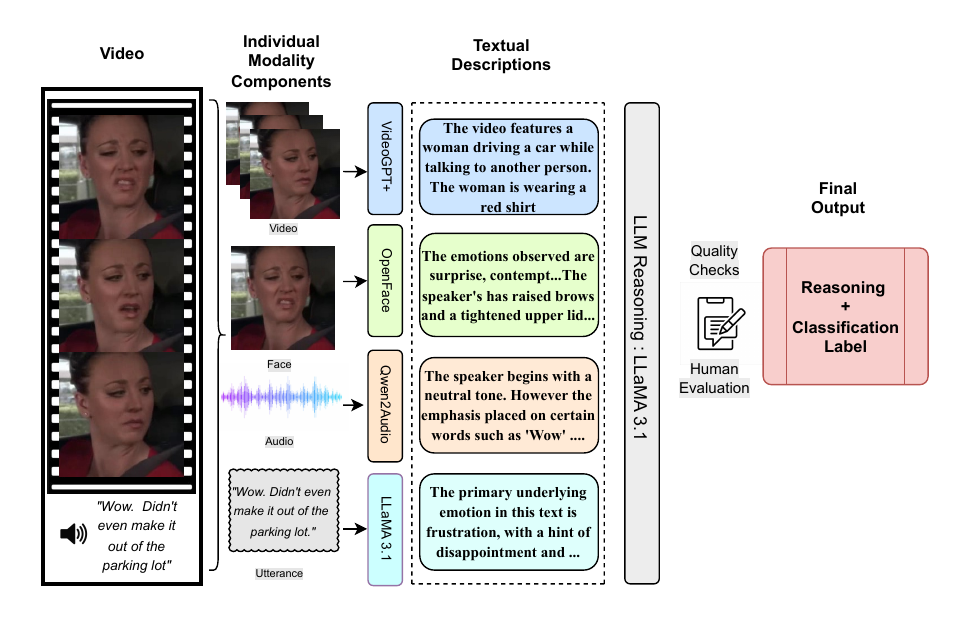} % Replace with your image path
    \caption{Sarcasm Reasoning Generation Pipeline for MUStReason}
    \label{fig:dataset}
\end{figure*}
\subsection{Video Understanding with LLMs}

% Video understanding involves dynamic multimodal inputs such as images, audio, and text with complex spatio-temporal dependencies important for tasks like activity recognition \cite{dhamsania2016survey} and behavior analysis \cite{borges2013video}. 
%%
VideoLMs encode sampled video frames and align the extracted visual features with corresponding text, which are then processed by large language models \cite{sun2019videobert,lin2024video,maaz2024videogptintegratingimagevideo,maaz-etal-2024-video,zhang2023video}.
Early models like VideoBERT \cite{sun2019videobert} were followed by later improvements in alignment using unified visual features in VideoLLaVA \cite{lin2024video}, segment-wise frame sampling \cite{maaz2024videogptintegratingimagevideo}, and Q-Former spatial-temporal modeling in Video-LLaMA \cite{zhang2023video}. Instruction-tuned models like VideoChatGPT \cite{maaz-etal-2024-video} enable conversational video understanding. 
However, a key limitation of most VideoLMs is the lack of audio processing \cite{cheng2024emotion}. 
This restricts their ability to capture audio cues which play an essential role in tasks like sarcasm detection. While recent models like VITA 1.5 \cite{fu2025vita} and Qwen2.5Omni \cite{xu2025qwen25omnitechnicalreport} have introduced audio encoding capabilities, VideoLMs continue to struggle with tasks that require higher-level reasoning \cite{li2025perception}. 
Our work focuses on enhancing sarcasm detection through pragmatic reasoning and identify gaps that hinder the ability of VideoLMs.
\subsection{Multimodal Reasoning}

Studies have explored how to elicit reasoning in Large Language Models (LLMs) for complex planning tasks \cite{huang2023towards}. 
CoT prompting has emerged as an effective technique to guide models in producing interpretable, stepwise reasoning chains, improving performance on challenging reasoning tasks \cite{10.5555/3600270.3601883,wei2022chain}.
In multimodal settings, early methods used a two-stage pipeline where vision models generated captions that LLMs then reasoned over \cite{gupta2023visual,yang2023mm}. 
Recent approaches instead prompt or train models directly on raw modality representations without verbalizing them \cite{li2025perception}. For example, \citet{zhang2023multimodal} combine textual and visual inputs to produce reasoning chains, while \citet{xu2025llava} propose staged reasoning in LLaVA-CoT.
Compositional CoT \cite{mitra2024compositional} leverages structured representations such as scene graphs to decompose visual scenes before reasoning, further improving interpretability and accuracy.
However, reasoning about sarcasm requires considering underlying pragmatics in addition to combining multimodal inputs. In the unimodal setting, LLMs have shown improved sarcasm classification performance when backed by pragmatic reasoning \cite{lee-etal-2025-pragmatic}. Driven by the success of staged reasoning and the necessity of pragmatic insights to interpret sarcasm, we introduce PragCoT, a pragmatic reasoning framework which processes multimodal cues.

\section{MUStReason}

Existing sarcasm detection datasets like MUStARD \cite{castro-etal-2019-towards} provide only coarse-grained binary labels and lack fine-grained annotations indicating which specific modalities or cues contribute to making the utterance sarcastic. Without explicit reasoning annotation, it is difficult to diagnose where and why models fail and localize whether the errors stem from wrong perception or from faulty inferential reasoning across the modalities. MUStReason bridges this gap by providing reasoning-aligned annotations that enable detailed evaluation of modality perception and inference failure which are key factors for assessing and improving pragmatic reasoning in multimodal models.
%This limits interpreting the ability of current VideoLMs to perform deeper, structured reasoning. To bridge this gap, we are the first to introduce MUStReason, a dataset having rich multimodal annotations that reveal how sarcastic or non-sarcastic interpretations are derived.

\subsection{Creation}

MUStReason builds on MUStARD++ Balanced \cite{bhosale2023sarcasmsightsoundbenchmarking}, which includes 691 sarcastic and 674 non-sarcastic video clips. We further add annotations that specify which attributes, individually or in combination, indicate sarcasm in each video. Figure \ref{fig:dataset}
illustrates our automatic process for generating annotations which decomposes sarcasm reasoning into two stages:\\

% \begin{enumerate}
%     \item \textbf{Perception of Independent Modalities}: Identify sarcasm-relevant cues from each modality using robust unimodal models.
%     \item \textbf{Pragmatic Reasoning Generation}: Synthesize these cues into a coherent reasoning.
% \end{enumerate}

\textbf{Stage 1: Perception of Independent Modalities}: We focus on perceiving sarcasm-relevant signals independently from  text, audio, and visual modalities. Drawing on insights from linguistic research \cite{cheang2008sound}, we select a targeted set of cues known to correlate with sarcastic expression, as outlined below.

\begin{itemize}
 \item\textbf{Video}: VideoGPT+ \cite{maaz2024videogptintegratingimagevideo} is used to describes the general scene, setting, and key characters or objects to provide contextual grounding.
 
 \item \textbf{Face}: We use OpenFace \cite{10.1109/FG.2018.00019} to identify facial muscle movements and expressions over time that reveal emotional states linked to sarcasm. 
 \item\textbf{Audio}: The audio is extracted from the corresponding video using PyDub, which is then passed to Qwen2-Audio \cite{chu2024qwen2audiotechnicalreport} to capture tonal attributes like mood, pitch changes, and stress patterns that influence the pragmatic meaning.  
 
 \item \textbf{Utterance}: Textual utterances (annotated
in the dataset) are passed to LLaMA 3.1 \cite{dubey2024llama} to analyze the emotional content of the spoken words and identify feelings such as mockery, contempt, or frustration.
 
 \end{itemize}
\begin{figure*}[t!]
    \centering    \includegraphics[width=0.9\linewidth]{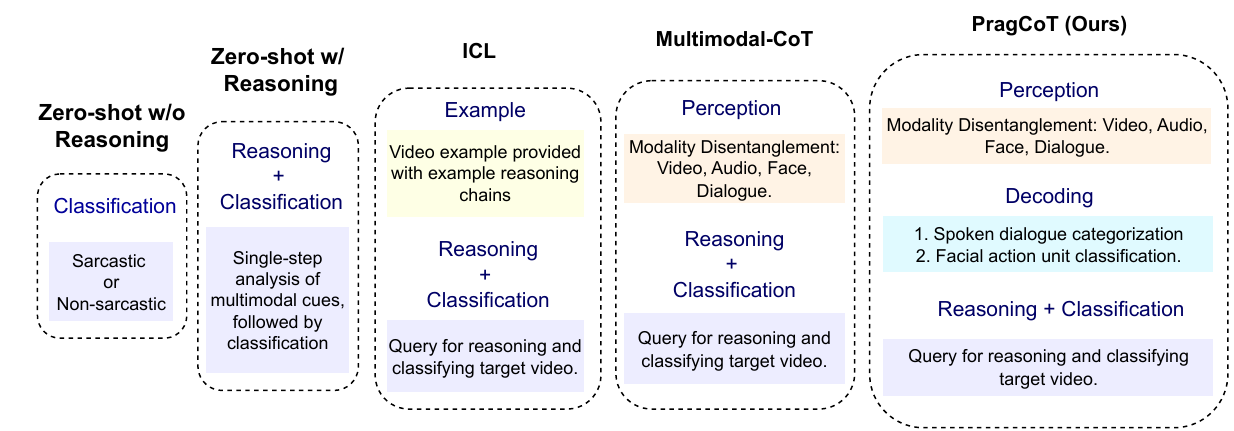} % Replace with your image path
    \caption{Comparison of prompting strategies for multimodal sarcasm reasoning. PragCoT extends standard CoT by explicitly decoding perceptual cues before reasoning and classification.}
    \label{fig:gradation}
\end{figure*}
\begin{figure}[t!] % Use [H] to place it exactly here, or [htbp] for floating
    \centering
    \includegraphics[width=0.6\linewidth]{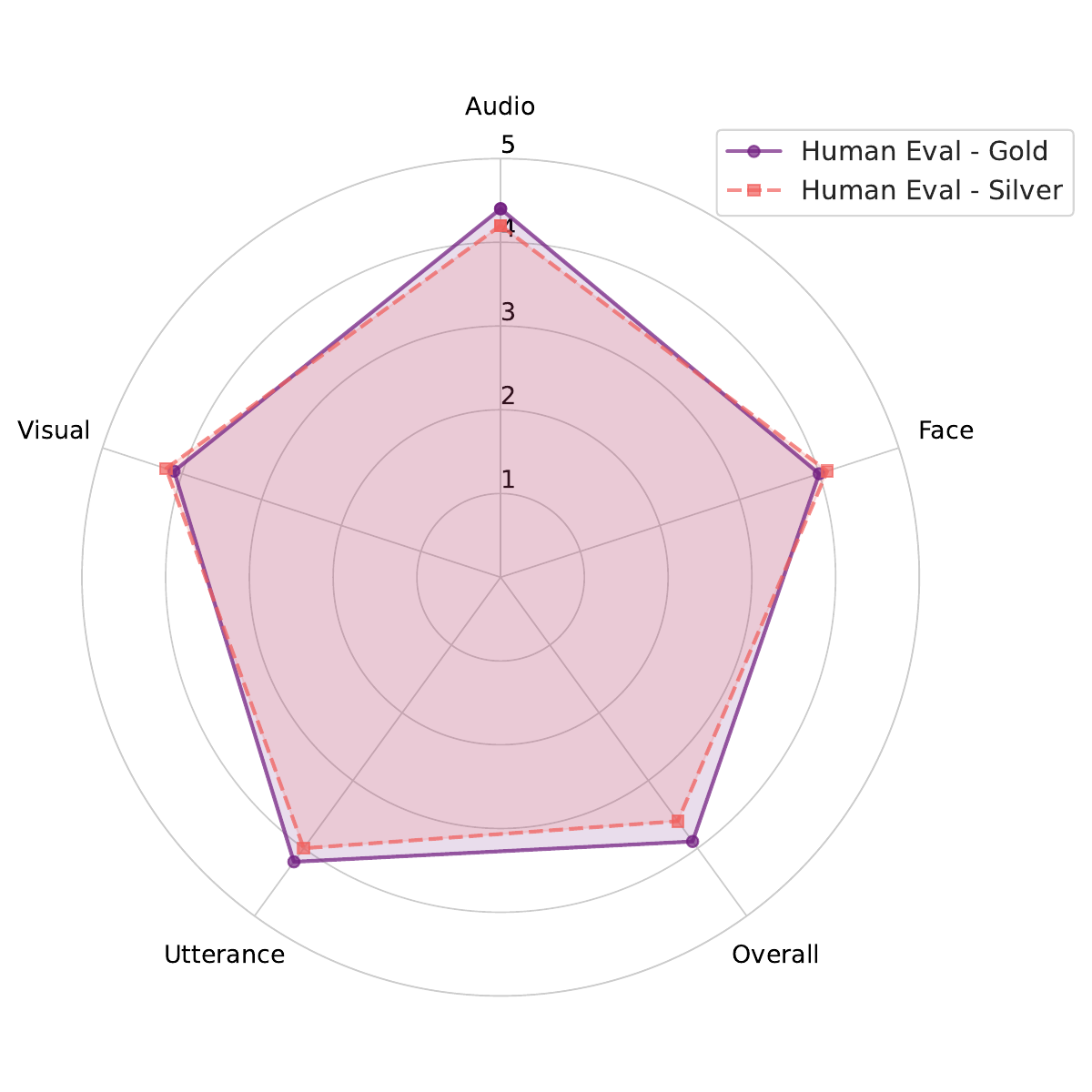} % Replace with your image file
    \caption{Human Evaluation of Annotation Quality in MUStReason.}
    \label{fig:quality-graph}
\end{figure}
\textbf{Stage 2: Pragmatic Reasoning Generation}: Once we obtain the individual modality descriptions, we prompt LLaMA 3.1 \cite{dubey2024llama} to collate these, and reason about why the co-occurence of various attributes or their presence in isolation results in the given label being sarcastic or non-sarcastic. The summary contains the most relevant cues which are instrumental in differentiating a sarcastic video from a non-sarcastic one.

\subsection{Evaluation of Generated Reasoning} 

To evaluate the quality of the automatically generated 
sarcasm reasoning, 
we curate a high-quality gold-standard set of 462 instances (approx. 34\% of the samples). These samples were checked by two annotators to ensure the correctness of individual modality descriptions, their relevance for sarcasm classification, and the coherence of the overall reasoning. Out of 462 silver instances converted to gold, 56\% of the samples required corrections, most of them pertaining to incorrect perception, especially for facial expressions and action units. The inter-annotator agreement (IAA) was calculated using Edit Distance (0.11) and BERTScore (0.99)\footnote{For comparison: Semantic similarity between annotation pairs for different videos are BERTScore: 0.88, Edit Distance: 0.65}, which ensured high lexical and semantic alignment between the annotators. In some cases there was a discrepancy between the model's final prediction and original label, and we excluded these samples. 

The remaining 66\% of the dataset comprises silver-standard samples (899), which underwent a sanity check to filter out irrelevant content. To estimate how different the silver data is from the gold data, we pass the gold annotations with their corresponding silver version to GPT-4.1 \cite{fu2024gptscore}, prompting it to evaluate the silver reasoning in terms of perceptual correctness and reasoning coherence compared to what is mentioned in the corresponding gold annotation. We find that the silver data achieves a rating of 4.51 out of 5 when compared to the gold data. To further validate the efficacy of silver data, we conduct a user study. We randomly select 10 samples each from the silver and gold sets and ask 26 participants to rate the reasoning on a 1–5 scale. Figure \ref{fig:quality-graph} shows the scores achieved by the silver and gold quality reasoning from GPT and human evaluators, supporting the overall quality of MUStReason\footnote{More details about dataset curation, annotation and prompts will be provided as part of the appendix.}.

\section{Experiments}
 To benchmark sarcasm detection in VideoLMs, we compare the following strategies with increasing order of guidance as shown in Figure \ref{fig:gradation}:\\
\textbf{Zero-Shot without Reasoning}: The prompt strictly asks the model to classify the video into either `sarcastic' or `non-sarcastic' class.\\
\textbf{Zero-Shot with Reasoning} \cite{han2023zero}: Given a video, the model is prompted to explain any sarcastic content and assign a label.\\
\textbf{In-Context Learning (ICL)} \cite{10.5555/3600270.3602070}: A video example along with its corresponding sarcasm reasoning is provided to the model. Given a target video, the model generates a reasoning including mention of the crucial attributes as in the reference reasoning. A practical limitation of this approach is that most VideoLMs are unable to process more than one video in a single conversation.\\
\textbf{Multimodal-CoT} \cite{zhang2023multimodal}:  A two-stage framework which fuses text and image information to generate a rationale, followed by inferential reasoning. We adapt this to our use case by first extracting individual modality descriptions and then collating them to perform reasoning.
\begin{table*}[t!]
    \centering
    \scriptsize
\begin{tabularx}{\textwidth}{p{4.8cm} | c | *{12}{X}} 
    \toprule
    \multirow{2}{*}{\textbf{Model}} & \multirow{2}{*}{\textbf{Modalities}} 
    & \multicolumn{2}{c}{\makecell{\textbf{ZS}\\\textbf{w/o Reasoning}}} 
    & \multicolumn{2}{c}{\makecell{\textbf{ZS}\\\textbf{w/ Reasoning}}} 
    & \multicolumn{2}{c}{\textbf{ICL}} 
    & \multicolumn{2}{c}{\makecell{\textbf{Multimodal}\\\textbf{CoT}}} 
    & \multicolumn{2}{c}{\makecell{\textbf{PragCoT}}} \\

    \cmidrule(lr){3-4} \cmidrule(lr){5-6} \cmidrule(lr){7-8} 
    \cmidrule(lr){9-10} \cmidrule(lr){11-12}

    & & Acc & F1 & Acc & F1 & Acc & F1 & Acc & F1 & Acc & F1 \\
    \midrule
    \textbf{Video-LLaVA} \cite{lin2024video} & T + V 
    & 49.6 & 33.9 & \textbf{52.2} & \textbf{42.9} & - & - & 50.3 & 33.7 & 50.2 & 34.1\\
    
    \textbf{VideoGPT+}  \cite{maaz2024videogptintegratingimagevideo} & T + V 
    & 49.7 & 35.0 & 50.5 & \textbf{39.8} & - & - & 50.7 & 33.6 & \textbf{56.4} & 36.1\\
    
    \textbf{ShareGPT4Video} \cite{chen2024sharegpt4video} & T + V 
    & 50.9 & 46.4 & 47.8 & 47.6 & - & -  & 54.9 & \textbf{54.1} & \textbf{55.0} & 52.8 \\
    
    \textbf{LLaVA-Next-Video}  \cite{zhang2024llavanextvideo} & T + V 
    & 50.2 & 38.5 & 49.6 & 35.7 & 50.0 & 33.3  & 52.3 & 48.4 & \textbf{54.8} & \textbf{53.2}\\
    
    \textbf{Qwen2.5VL} \cite{Qwen2.5-VL} & T + V 
    & 49.6 & 39.5 & 50.6 & 35.4 & 47.8 & 32.3  & \textbf{58.0} & \textbf{57.2} & 57.4 & 55.3\\
    
    \midrule
    \textbf{VITA} \cite{fu2025vita} & T + V + A 
    & 49.6 & 33.7 & 49.5 & \textbf{33.6} & - & -  & 49.4 & 33.2 & \textbf{49.6} & 33.4\\
    
    \textbf{Qwen2.5Omni} \cite{xu2025qwen25omnitechnicalreport} & T + V + A 
    & 49.6 & 39.5 & 48.9 & 34.3 & 55.0 & 42.2  & 57.0 & 56.9 & \textbf{59.5} & \textbf{59.3}\\
    \bottomrule
\end{tabularx}
    \caption{Performance of Video-Language Models on Sarcasm Classification using CoT Prompting (Acc: Accuracy, F1: macro F1-Score).}
    \label{tab:cot}
\end{table*}

% \begin{table}[t!]
% \centering
% \begin{tabularx}{\columnwidth}{l|cc}
% \toprule
% \textbf{Components} & \textbf{Acc} & \textbf{F1} \\
% \midrule
% PragCoT \textit{w/o  Dialogue Decoding} & 56.6 & 56.6 \\
% PragCoT \textit{w/o Face Decoding} & 59.0 & 58.7 \\
% PragCoT & \textbf{59.5} & \textbf{59.3} \\
% \bottomrule
% \end{tabularx}
% \caption{Ablation of the Decoding step in PragCoT for Qwen2.5Omni. Addition of the Decoding step shows improvement in classification performance.}
% \label{tab:ablation}
% \end{table}
\begin{table}[t!]
\centering
\resizebox{0.8\columnwidth}{!}{
\begin{tabular}{l|cc}
\toprule
\textbf{Components} & \textbf{Acc} & \textbf{F1} \\
\midrule
PragCoT \textit{w/o Dialogue Decoding} & 56.6 & 56.6 \\
PragCoT \textit{w/o Face Decoding} & 59.0 & 58.7 \\
PragCoT & \textbf{59.5} & \textbf{59.3} \\
\bottomrule
\end{tabular}
}
\caption{Ablation of the Decoding step in PragCoT for Qwen2.5Omni.}
\label{tab:ablation}
\end{table}
\textbf{PragCoT (Ours)}: To enable VideoLMs to reason about sarcasm pragmatically, we propose a structured prompting approach called PragCoT, a variant of Zero-Shot CoT prompting \cite{10.5555/3600270.3601883}. Developing complex reasoning abilities through the integration of information from multiple modalities and the systematic resolution of sub-problems has proven highly effective in multimodal contexts \cite{zhang2023multimodal, feivideo}. However, while multimodal reasoning draws inferences from the combined content of modalities, pragmatic reasoning contains an additional layer of complexity:  It relies heavily on contextual cues, social dynamics and the viewer's interpretation of a situation, focusing more on the underlying intent rather than the literal meaning. PragCoT bridges this gap by introducing a modality-decoding step in addition to the perception and reasoning steps: 

\begin{enumerate}
    \item \textbf{Perception}: The model is queried for speaker utterance, acoustic features, generic facial attributes and video backdrop context, separately to disentangle information from the modalities present. 
    \item \textbf{Decoding}: Pragmatic reasoning unlike traditional reasoning aims to derive non-literal meaning conveyed by a situation based on speaker intent and interaction \cite{grice1975logic}. This interpretation heavily relies on how a dialogue is framed \cite{goffman1981forms} especially in conjunction with facial attributes, the most immediate visual evidence. For instance, when the content of an utterance is not congruent with how it is delivered (e.g., saying \textit{"Thank You!"} with a deadpan face), the listener's ability to decode the sarcastic intent is significantly enhanced compared to relying on only words or tone \cite{article}. This step, thus, facilitates gathering specific ingredients required for pragmatic reasoning. To capture minute facial attributes, we prompt the model to classify which facial action units from a provided list are present on the speaker's face. Further, if sarcasm is self-contained in a dialogue, it can render a video sarcastic irrespective of visual or auditory cues. Hence, we ask the model to categorize dialogues into one of the following classes: neutral, metaphoric, ironic or hyperbolic \cite{burgers2016figurative}.
    \item \textbf{Reasoning}: Finally, the model is given detailed instructions about the kind of cues it should attend to, disentangle the incongruity among intents, generate a rationale and aggregate its analysis into a comprehensive reasoning, consisting of the classification label.   
\end{enumerate} 

%as spoken words are a key source of sarcasm cues.
% \begin{figure}[t!]
%     \centering
%     \includegraphics[width=\linewidth]{images/cotv4.pdf} % Replace with your image path
%     \caption{The PragCoT Framework showing the steps for Sarcasm Identification - $\textit{Perception}$: Where each individual modality is queried for, along with intent. $\textit{Reasoning}$: The cues from the Perception step are combined to reason about the presence of sarcasm in the video.}
%     \label{fig:pragcot}
% \end{figure}

% \subsection{Video Understanding with LLMs}
For the model choices we use off-the-shelf Video\-LMs from the leaderboard \cite{fu2025video}. Among the 7 models that we evaluate, 5 of them do not support audio. We provide these models explicitly with the transcriptions of the speaker utterances, which are available in the original dataset.
\subsection{Evaluation Metrics}
Our evaluation framework assesses performance of VideoLMs with respect to sarcasm classification and quality of generated reasoning. Following previous work \cite{castro-etal-2019-towards, zhang2021multi}, accuracy and macro-F1 are chosen as metrics for binary classification. Next, we leverage the detailed annotations in MUStReason to assess the perceptual correctness and reasoning coherence of the sarcasm explanations generated by VideoLMs. We conduct an automated GPT4.1 \cite{achiam2023gpt} based evaluation to analyze whether the model explanations align with the annotated reasoning in MUStReason on the basis of correctness of input modality attributes (e.g., spoken words, facial expressions, audio tone), reasoning, and overall understanding of sarcasm. 

\begin{table*}[t!]
\centering
\scriptsize
\begin{tabularx} {\textwidth}{|X|}
\hline
% \multicolumn{2}{|c|}{
\multicolumn{1}{|c|}{
    \includegraphics[width=1.8cm]{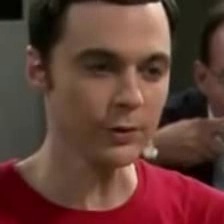} 
    \includegraphics[width=1.8cm]{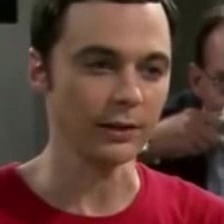}
    \includegraphics[width=1.8cm]{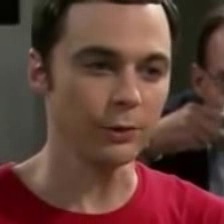}
    \includegraphics[width=1.8cm]{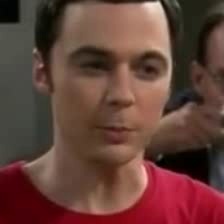}
    \includegraphics[width=1.8cm]{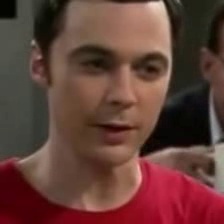}
    \includegraphics[width=1.8cm]{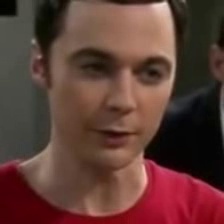}
    \includegraphics[width=1.8cm]{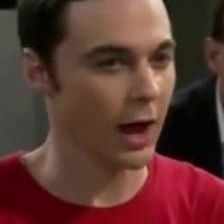}
    \includegraphics[width=1.8cm]{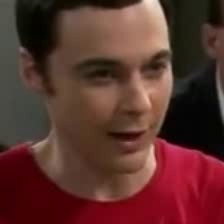}
} \\
\multicolumn{1}{|c|}{\textit{\textbf{Utterance}: ``Interesting. You see betrayal in others, but not yourself." }} \\
\hline
\hline
\rule{0pt}{3ex} 
\textbf{Gold: } The video is sarcastic because of the speaker's utterance ``you see betrayal in others, but not yourself", which implies hypocrisy and double standards, conveying a negative sentiment with a hint of frustration and anger. The tone is assertive and slightly confrontational, and the facial expressions, such as raised brows and forced smile, supports the sarcastic tone. The irony and criticism in the statement, combined with the speaker's composed and unemotional delivery, also contribute to the sarcastic label.
% \rule{0pt}{1ex}
\\
\hline

\rule{0pt}{3ex} 
\textbf{ZS w/ Reasoning:} To analyze the video for sarcasm, let's break down the elements: The setting appears to be a casual dining area, possibly a cafeteria or a similar communal space. There are other people around, some eating and others engaged in conversation. the environment seems relaxed and informal. \textcolor{blue}{The audio tone is not explicitly provided}, but we can infer from the visual context that the speaker might be using a neutral or slightly casual tone, as \textcolor{red}{there are no indications of an exaggerated or mismatched delivery that would suggest sarcasm}.The main character is seen speaking while holding a fork and a drink. \textcolor{red} {His facial expression does not show any obvious signs of sarcasm such as raised eyebrows or a forced smile}. instead, he appears to be engaged in a normal conversation. \textcolor{blue}{The dialogue is not provided in the description, so we cannot determine the sentiment directly}. However, based on the context, it seems to be a neutral conversation without any indication of sarcasm. Given the lack of any clear indicators of sarcasm in the audio tone, facial expressions, and dialogue sentiment, \textcolor{red}{the scene does not provide strong evidence for sarcasm.}
\rule{0pt}{1ex}
\\
\hline
% \\
% \rule{0pt}{3ex} 
\rule{0pt}{3ex} 
\textbf{ICL:} The video is sarcastic. The speaker's tone is likely to be mocking or dismissive, which is a common characteristic of sarcasm. The dialogue sentiment is negative, as the speaker is pointing out a perceived flaw in the other person's ability to recognize their own betrayal. The facial expressions and AUs, such as raised eyebrows or a forced smile, would likely contradict the spoken words, further indicating sarcasm. \textcolor{blue}{The setting appears to be a casual environment, possibly a workplace or social gathering, which could contribute to the context of the sarcastic remark.} 
% \rule{0pt}{1ex}
\\
\hline
\rule{0pt}{3ex} 
\textbf{Multimodal-CoT:} The video segment is likely to be ironic or sarcastic. The serious tone and content of the speech, combined with the incongruity between the serious tone and the casual setting, suggest that the speaker is making a pointed observation about the other person's behavior \textcolor{olive}{in a way that is intended to be taken seriously but may also be interpreted as ironic or sarcastic}.

% \rule{0pt}{1ex}
\\
\hline
% \\
% \rule{0pt}{1ex}
\rule{0pt}{3ex} 
\textbf{PragCoT:} The video segment is sarcastic. The utterance cues, particularly the use of ``interesting" and the direct contradiction in the statement, clearly indicate sarcasm. The audio cues support this with an accusatory and slightly dismissive tone, as well as a slightly raised pitch at the end of the sentence. The facial expression cues, while neutral to slightly concerned, do not negate the sarcastic tone of the speech. The overall interaction between the tone and the facial expression further reinforces the sarcastic nature of the statement.\\
\hline

\end{tabularx}

\caption{Example of sarcasm reasoning generated by Qwen2.5Omni model for different prompting strategies. The texts marked in \textcolor{red}{red}, \textcolor{blue}{blue} and \textcolor{olive}{olive} indicate wrong attribute prediction, hallucination and uncertainty, respectively.}
\label{tab:examples}
\end{table*}

\section{Results and Analysis}

MUStReason serves dual purpose:  (1) It acts as a benchmark to help evaluate the reasoning quality of VideoLMs for the task of sarcasm detection. (2) The detailed, modality-focused annotations facilitate the identification of failure cases, reveal whether VideoLMs attend to the relevant attributes, and highlight gaps in perception and reasoning.

\subsection{Benchmarking Sarcasm Detection Performance of VideoLMs}
Table \ref{tab:cot} shows classification results (average of 3  runs) across 5 methods for 7 VideoLMs. PragCoT resulted in higher F1 scores with statistically significant (p-value$=$0.025 $<$ 0.05) improvements of up to 20\% compared to zero-shot classification, about 17\% with respect to ICL and 2.4\% compared to Multimodal-CoT. We observe that VideoLMs perform quite poorly in the zero-shot classification setting. In fact for few models (Qwen2.5VL, LLaVA-Next-Video and Qwen2.5Omni), the F1-score goes down when the model generated an accompanying reasoning compared to zero-shot classification without reasoning. Low performance in the zero-shot setting suggests that the models might not be attending to the appropriate cues during classification. Alternatively, it could indicate the model's dependence on implicit heuristics for sarcasm detection, with a limited ability to articulate intermediate reasoning \cite{anantha-ramakrishnan-etal-2025-cordial}. ICL shows slight performance gains over zero-shot classification for the Qwen2.5Omni model only, indicating that providing an example might be helpul but not entirely reliable. Multimodal-CoT, owing to its multi-step reasoning process achieves higher scores than the single-step methods. PragCoT achieves best performance for majority of the models highlighting how crucial the decoding step is for better perception-reasoning alignment. We provide evidence of empirical improvement on addition of the decoding step in Table \ref{tab:ablation}.
%We observe that PragCoT achieves significant improvement than other approaches by ensuring that the model incorporates insights from each individual modality, rather than constructing a reasoning chain that overlooks important cues.

\begin{figure}[t!]
    \centering
    \includegraphics[width=0.6\linewidth]{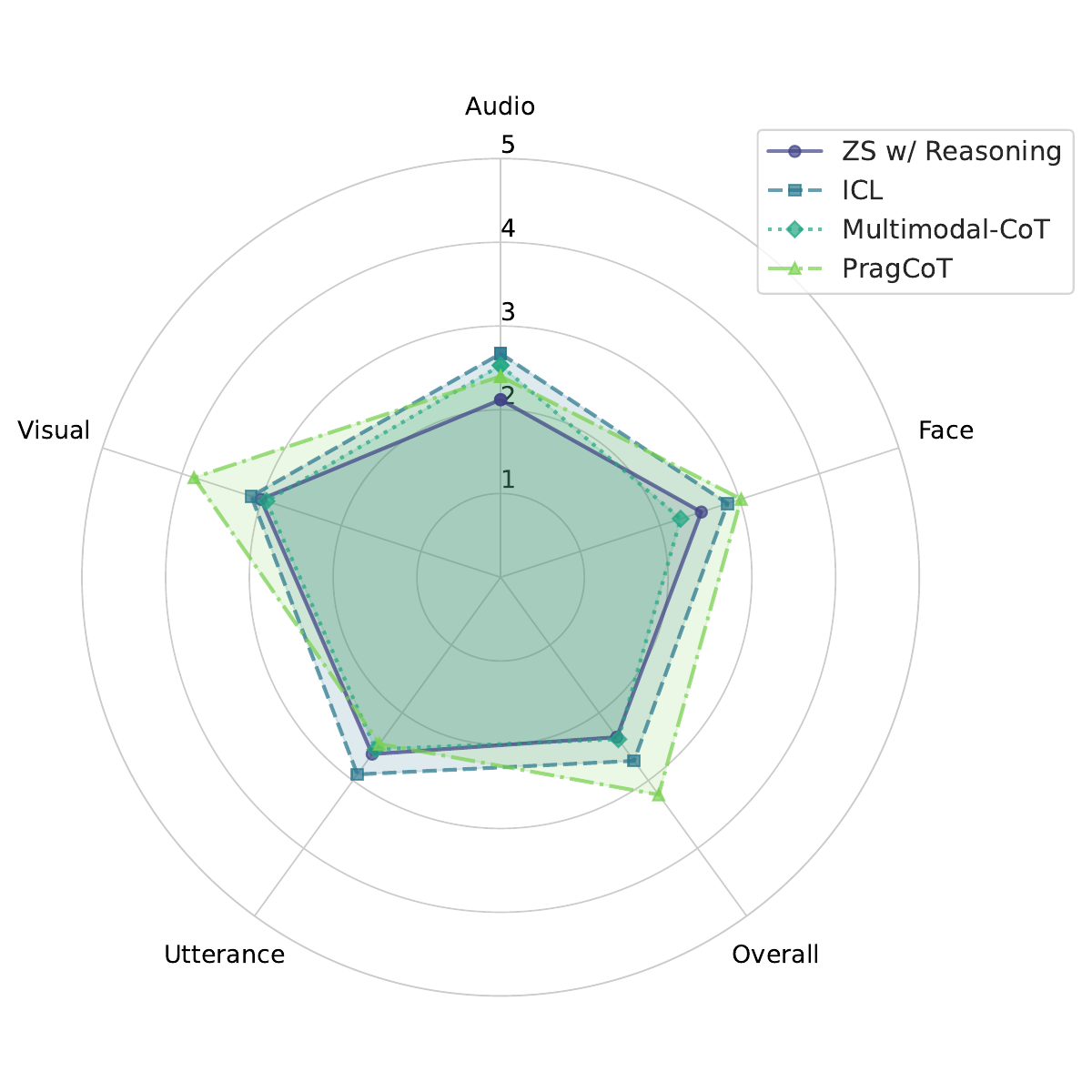} % Replace with your image path
    \caption{Qualitative Evaluation of Model Generated Reasoning.}
    \label{fig:spider}
\end{figure}

\textbf{Qualitative Evaluation of Reasoning: }
In addition to assessing sarcasm classification performance, we also evaluate the generated reasoning using MUStReason by comparing the generated reasoning with the annotated data. GPT4.1 judges the sarcasm reasoning based on a set of guidelines, ensuring the credibility of the perceptual cures and the overall relevance of the reasoning. Figure \ref{fig:spider} displays the ratings provided by GPT for the overall quality of the reasoning and description of individual modalities within the reasoning generated by the discussed prompting techniques compared to the gold reasoning. We observe a consistent improvement across most aspects of the reasoning generated via PragCoT. The visual modality descriptions receive the highest ratings, likely because VideoLMs are predominantly trained on scene understanding. In contrast, they show limited ability in capturing intent and audio tone.

Table \ref{tab:examples} presents examples of sarcasm reasoning generated by different prompting techniques for the Qwen2.5Omni model. 
% We observe that zero-shot reasoning is prone to hallucinations. Even though Qwen2.5Omni can interpret audio, the reasoning mentions lack of any audio input. 
We observe that zero-shot reasoning is prone to hallucinations; although Qwen2.5Omni can interpret audio, its reasoning often claims the absence of any audio input. ICL vividly articulates the factors contributing to sarcasm but tends to include irrelevant factual arguments. This suggests that exposure to examples helps teaches the model to perceive the essential cues, but it still fails to build meaningful association between them. 
Multimodal-CoT incorporates standard perceptual cues such as speech content but often overlooks finer indicators like facial expressions, that hinders coherent reasoning as highlighted in the example. In contrast, PragCoT integrates essential cues, including facial expressions, achieving more accurate sarcasm detection by distinguishing between literal and implied meanings.
% For Multimodal-CoT, the perceptual cues are collated but the reasoning follows an uncertain chain-of-thought.
% PragCoT summarizes all the crucial details and is more successful at recognizing sarcasm, thanks to distinguishing between literal and implied meaning.

\subsection{Analyzing Perception and Reasoning in VideoLMs for Determining Sarcasm}
\begin{figure*}[t!]
    \centering    \includegraphics[width=.9\linewidth, clip=TRUE, trim=0cm 2.5cm 0cm 2cm]{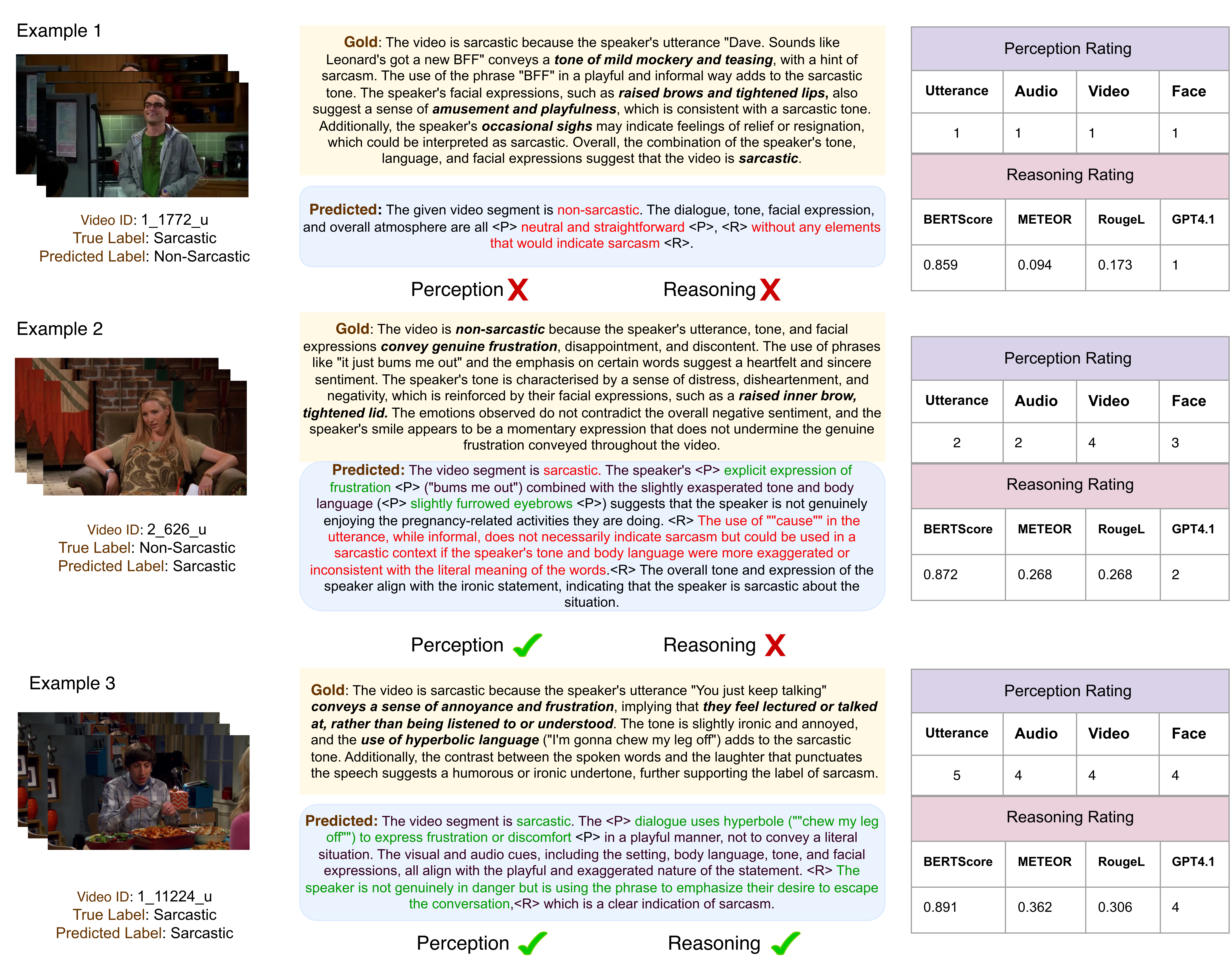} % Replace with your image path
    \caption{Qualitative and quantitative analysis of model-generated reasoning using annotations from the MUStReason dataset. These annotations help in identifying whether the model fails in perception, reasoning or both when identifying sarcasm. \textbf{Bold} words in Gold indicate the attributes present in Predicted. Text marked in \textcolor{green}{green} and \textcolor{red}{red} indicate correct and wrong predictions respectively. Tags $<$P$>$ and $<$R$>$ represent perception and reasoning respectively.}
    \label{fig:analysis}
\end{figure*}

We observed that even though classification preceeded by reasoning improves accuracy, VideoLMs perform quite poorly in classifying sarcasm compared to state-of-the-art models \cite{10887877} fine-tuned for the task. VideoLMs are composed of encoders which process image frames, speech and text, followed by an LLM decoder capable of reasoning and decision-making. This leads us to examine whether these models fail in accurately perceiving multimodal cues essential for detecting sarcasm or struggles to construct the reasoning chain. MUStReason lays a ground for investigation by providing gold-standard sarcasm reasoning annotations, which could be qualitatively and quantitatively compared with model-generated responses to identify the loopholes. Upon analyzing the failure cases, we identify two types of errors as observed in Figure \ref{fig:analysis}: \\
\textbf{Failure in both perception and reasoning}:
 In Example 1, we observe that the model vaguely assigns a \textit{neutral} tag to all the modalities, without describing what each modality individually conveys. Lack of concrete details results in minimal reasoning by the model. Comparing with the gold annotation it is evident how the mocking utterance, in conjunction with the speaker's expressions and actions, supports the conclusion that the video is sarcastic. This highlights the importance of accurate perception as a prerequisite for reasoning when tackling complex tasks like sarcasm detection. Thus, if the model fails to attend to key perceptual cues, the reasoning thread gets disrupted, resulting in an incorrect prediction.\\
\textbf{Failure in reasoning with correct perception}:
 For Example 2, the model correctly identifies \textit{expression of frustration} and \textit{furrowed eyebrows} of the speaker. The gold annotation shows how reasoning is grounded in these attributes. In contrast, a preceding spurious cue generated by the model subsequently leads to an under-confident reasoning chain.  It completely overlooks the correct cues it had identified earlier, leading to an incorrect prediction. In this type of error, even though the model correctly perceives information from the different modalities, it lacks confidence in the reasoning process or fails to infer the association between the cues that would lead to the correct label.

We further leverage the MUStReason dataset to evaluate the model's reasoning quantitatively. Firstly, we compare the model's response with the gold reasoning using semantic metrics like BERTScore \cite{zhang2019bertscore}, METEOR \cite{banerjee2005meteor} and RougeL \cite{lin2004rouge} and observe that the inclusion and exclusion of cues and reasoning steps reflects in the scores. Since reasoning might have low semantic similarity while still conveying the same meaning, we use GPT as a judge  to compare the model generated response with gold annotation. The evaluation rates the presence of individual modality descriptions and reasoning in a scale of 1-5,  from highly dissimilar to highly similar. For Example 1, which falls short in both perception and reasoning, the scores are 1, further establishing the reason of failure. In Example 2, while there is some similarity in video context and facial expressions, the overall rating remains low due to incorrect construction of reasoning chain. To examine how reasoning quality degrades from correct to incorrectly classified samples, we calculate the mean of the ratings provided by GPT for overall reasoning quality for each class, obtaining scores of 3.02 for correct and 1.64 for incorrect predictions, on a 5-point scale.

\begin{table}[t!]
\centering

\scriptsize
\begin{tabularx}{0.9\columnwidth}{lXX}
\toprule
\textbf{Model} & \textbf{Acc} & \textbf{F1} \\
\midrule
\multicolumn{3}{l}{\textit{Large Language Models (LLMs)}} \\
\midrule
\textbf{Llama-3.1-8B-Instruct} \cite{dubey2024llama} & 59.0 & 57.6 \\
\textbf{Qwen3LM-8B} \cite{yang2025qwen3}& 60.5 & 60.0 \\
\textbf{Mistral-7B-Instruct-v0.3} \cite{jiang2023mistral7b} & 61.8 & 61.7 \\
\textbf{GPT-5} & \textbf{65.6} & \textbf{65.6} \\
\midrule
\multicolumn{3}{l}{\textit{Video-Language Models (VideoLMs)}} \\
\midrule
\textbf{LLaVA-Next-Video}  & 60.6 & 60.4 \\
\textbf{Qwen2.5VL}  & \textbf{61.8} & \textbf{61.7} \\
\textbf{Qwen2.5Omni}  & 60.7 & 59.7 \\

\bottomrule
\end{tabularx}
\caption{Sarcasm classification performance of LLMs vs VideoLMs when provided with quasi-perfect perception.}
\label{tab:multicot}
\end{table}

% \begin{enumerate}
%     \item From Human Eval 1: Investigate why the ratings are low for the annotated data. Check the prediction/reasoning for the same sample for model-generated response. Is it right/wrong? Why so? Does the VideoLM also generate poor quality reasoning? Why so? Similar analysis for video having high ratings for each.
%     \item From Human Eval 1: Cases where individual descriptions are good but fails in overall quality --> fails to build the association/reasoning.
%     \item From Human Eval 1: One modality failing translates to wrong reasoning case.
%     \item Modality ablations? (Check bootstrapping results before adding)
% \end{enumerate}
% \begin{figure}
%     \centering
%     \includegraphics[width=\linewidth]{latex/images/Human_exp_1.png}
%     \caption{Human Eval 1}
%     \label{fig:placeholder}
% \end{figure}
% \anisha{Deductions from table 2} \\

\textbf{Do LLMs outperform VideoLMs in sarcasm detection when provided with quasi-perfect perceptual cues?}

During error analysis we observe that incorrect perception misguides the model to build an incorrect reasoning chain. This raises the question: given near-perfect modality descriptions derived individually from \textit{modality} $\rightarrow$ \textit{text} models, how well can the VideoLMs reason? To investigate this, we undertake an approach where the model is supplied with individual modality descriptions and tasked to only reason and explain. This is typically employed for LLMs that lack the ability to process audio or visual inputs.  In our setup, we extend this method to VideoLMs as well. This helps in assessing the reasoning ability of these models. Table \ref{tab:multicot} shows a marginal increase in classification scores for Qwen2.5Omni compared to Table \ref{tab:cot} indicating the model suffers from a reasoning bottleneck even when perception is improved. For LLaVA-Next-Video and Qwen2.5VL, the increase is higher, suggesting a substantial gap in perception as well.  This proves that finer-grained perception is an essential first step to reason about complex pragmatic phenomenon like sarcasm. LLMs achieve performance comparable to VideoLMs, with GPT-5 being the best reasoner, beating the open-source LLMs. However, the overall classification performance remains low for both LLMs and VideoLMs, indicating that these models are still far from demonstrating pragmatic reasoning capabilities.

\section{Conclusion}
In his work, we analyze the performance gaps of existing open-source VideoLMs in perceiving, reasoning and detecting sarcasm, which requires interpreting and disambiguating conflicting information gathered from diverse modalities, we introduce the MUStReason diagnostic benchmark which contains samples annotated with sarcasm reasoning. Furthermore, we device a multi-stage structured prompting approach, PragCoT, in order to incorporate pragmatic reasoning abilities in VideoLMs, which is key to understanding and inferring underlying sarcastic implications in videos. While PragCoT provides one example of how MUStReason can be used to evaluate reasoning behavior, the annotations can support evaluation of any generic reasoning framework and similar reasoning annotations can be generated to assess tasks like humor detection and figurative language understanding. 

\section{Limitations}
A primary limitation of our work lies in the sarcasm reasoning annotations being derived from the  MUStARD++ Balanced dataset. Thus, the contextual cues and non-verbal markers present in the annotations are dataset-specific, preventing generalizability across dyadic conversations which might contain additional cue variations not represented in these annotations. Besides, the dataset is in English. Since sarcasm is influenced by cultural and social dynamics, the monolingual nature of the dataset misses language-specific sarcasm indicators. In addition, while PragCoT lays the foundation for pragmatic reasoning for sarcasm interpretation by capturing contradictions and incongruities between modalities, integrating ensemble approaches or reinforcement learning based techniques into this framework might help these models generate better reasoning chains. Currently the diagnostic benchmark is limited to sarcasm. Future work would focus on extending it to other tasks like irony and figurative language understanding to make pragmatic reasoning in multimodal models interpretable.

\section{References}\label{sec:reference}
\bibliographystyle{lrec2026-natbib}
\bibliography{custom}

% \subsection{Language Resource References}
% \label{lr:ref}
% \bibliographystylelanguageresource{lrec2026-natbib}
% \bibliographylanguageresource{languageresource}

\end{document}